\documentclass[letterpaper]{article} % DO NOT CHANGE THIS
\usepackage{aaai24}  % DO NOT CHANGE THIS
\usepackage{times}  % DO NOT CHANGE THIS
\usepackage{helvet}  % DO NOT CHANGE THIS
\usepackage{courier}  % DO NOT CHANGE THIS
\usepackage[hyphens]{url}  % DO NOT CHANGE THIS
\usepackage{graphicx} % DO NOT CHANGE THIS
\usepackage{natbib}  % DO NOT CHANGE THIS AND DO NOT ADD ANY OPTIONS TO IT
\usepackage{caption} % DO NOT CHANGE THIS AND DO NOT ADD ANY OPTIONS TO IT
\usepackage{algorithm}
\usepackage{algorithmic}
\usepackage{amsmath}
\usepackage{amsthm}
\usepackage{amssymb}
\usepackage{booktabs}
\usepackage{bbding}
\usepackage{xspace} 
\usepackage{newfloat}
\usepackage{listings}
\DeclareCaptionStyle{ruled}{labelfont=normalfont,labelsep=colon,strut=off} % DO NOT CHANGE THIS
\floatstyle{ruled}
\newfloat{listing}{tb}{lst}{}
\floatname{listing}{Listing}
\newcommand{\eat}[1]{}
\newcommand{\paratitle}[1]{\vspace{1ex}\noindent \textbf{#1}}

\let\oldhat\hat
\renewcommand{\vec}[1]{\mathbf{#1}}
\renewcommand{\hat}[1]{\oldhat{\mathbf{#1}}}
\renewcommand{\matrix}[1]{\mathbf{#1}}
\newcommand{\eg}{\emph{e.g.,}\xspace}

\newcommand{\ie}{\emph{i.e.,}\xspace}
\newcommand{\etc}{\emph{etc.}\xspace}

\title{Detection-based Intermediate Supervision for Visual Question Answering}
\author {
    Yuhang Liu\equalcontrib\textsuperscript{\rm 1,\rm 2,\rm 4},
    Daowan Peng\equalcontrib\textsuperscript{\rm 1,\rm 2},
    Wei Wei\thanks{Corresponding author.}\textsuperscript{\rm 1,\rm 2},
    Yuanyuan Fu\textsuperscript{\rm 2,\rm 3},
    Wenfeng Xie\textsuperscript{\rm 2,\rm 3},
    Dangyang Chen\textsuperscript{\rm 2,\rm 3}
}
\affiliations {
    \textsuperscript{\rm 1} CCIIP Lab, School of Computer Science and Technology, Huazhong University of Science and Technology\\
    \textsuperscript{\rm 2}Joint Laboratory of HUST and Pingan Property \& Casualty Research (HPL)\\
    \textsuperscript{\rm 3}Ping An Property \& Casualty Insurance Company of China, Ltd.\\
    \textsuperscript{\rm 4}ByteDance Inc.\\
    
    liuyuhang.ysy@bytedance.com, \{pengdw, weiw\}@hust.edu.cn, fuyuanyuan83@gmail.com, \{xiewenfeng801, chendangyang273\}@pingan.com.cn
}
\usepackage{bibentry}
\begin{document}

\maketitle

\begin{abstract}
Recently, neural module networks (NMNs) have yielded ongoing success in answering compositional visual questions, especially those involving multi-hop visual and logical reasoning. NMNs decompose the complex question into several sub-tasks using instance-modules from the reasoning paths of that question and then exploit intermediate supervisions to guide answer prediction, thereby improving inference interpretability. 
However, their performance may be hindered due to sketchy modeling of intermediate supervisions. For instance, (1) a prior assumption that each instance-module refers to only one grounded object yet overlooks other potentially associated grounded objects, impeding full cross-modal alignment learning; (2) IoU-based intermediate supervisions may introduce noise signals as the bounding box overlap issue might guide the model's focus towards irrelevant objects. 
To address these issues, a novel method,  \textbf{\underline{D}}etection-based \textbf{\underline{I}}ntermediate \textbf{\underline{S}}upervision (DIS), is proposed, which adopts a generative detection framework to facilitate multiple grounding supervisions via sequence generation. As such, DIS offers more comprehensive and accurate intermediate supervisions, thereby boosting answer prediction performance. Furthermore, by considering intermediate results, DIS enhances the consistency in answering compositional questions and their sub-questions. 
Extensive experiments demonstrate the superiority of our proposed DIS, showcasing both improved accuracy and state-of-the-art reasoning consistency compared to prior approaches.
% The codes are in the supplementary material and will be open-sourced to facilitate future research.
\end{abstract}

% \vspace{-0.4cm}
\section{Introduction}

Compositional visual question answering (VQA) \cite{Hudson2019GQAAN,Chen2019MetaMN,Jing2022MaintainingRC} has been an emerging research topic in multimodal domain, and received increasing attention from both computer vision and the language processing communities. Specifically, given input images and questions, it is required to generate answers for the questions according the content of images. Generally, compositional questions involve multiple visual entities or concepts (\ie objects, attributes, and relations), and the models demand a rich set of abilities (semantic understanding, object detections, visual/logical reasoning) to get the right answer. 
One of the challenges in compositional visual question answering lies in modeling the reasoning process. From this perspective, VQA models can be divided into two categories, \ie holistic and modular. Holistic models \cite{Hu2019LanguageConditionedGN,Yu2019DeepMC,Yang2020TRRNetTR} generate answers for all types of questions through a unified multimodal fusion model, and the reasoning process is implicitly performed during the encoding and fusion stages. Despite the effectiveness, holistic models cannot reflect the intermediate reasoning process. On the contrary, modular model, \ie neural module network (NMN) \cite{Andreas2015NeuralMN,Hu2018ExplainableNC,Hudson2018CompositionalAN,Chen2019MetaMN}, has been a mainstream approach due to its explicit reasoning procedure and interpretable characteristics. Specifically, NMN parses questions into predefined reasoning modules and composes these modules into an executive program, thus deconstructing complex questions into several easy-to-solve problems. During the process of multi-hop visual/logical reasoning and answering generation, the intermediate status and results can be explicitly reflected from each module. Moreover, in order to restrict the reasoning process of NMN models, extra intermediate supervisions \cite{Chen2019MetaMN,Zhao2021ProToPT} are proposed to improve the answer prediction performance, which restricts models to focus on pivotal objects via Intersection over Union (IoU) constraint between predicted bounding boxes and ground truth ones.

\begin{figure}[t]
    \centering
    \includegraphics[width=\linewidth]{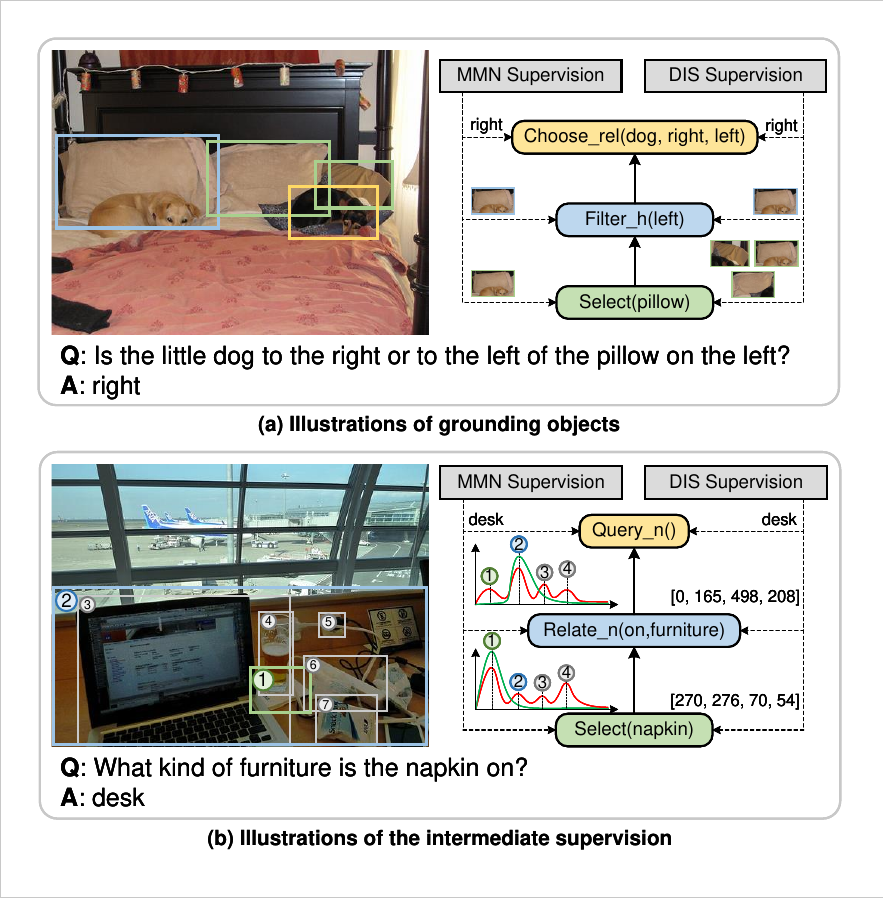}
    \caption{Illustrations of the differences in intermediate supervisions between MMN and DIS: (a) Number of grounded objects, and (b) IoU-based and detection-based supervisions (green and red lines indicate the ground-truth and IoU-based probability distributions over objects).}
    \label{fig:sample}
    \vspace{-0.5cm}
\end{figure}

Despite the significant improvement in the accuracy metric, there are several shortcomings in the IoU-based intermediate supervisions of previous methods. For example, MMN \cite{Chen2019MetaMN} directly exploits the ground-truth scene graph paths corresponding to questions for intermediate result generation, which ignores other potentially correct intermediate objects and hinders the model to fully learn cross-modal alignment and reasoning process. As Figure \ref{fig:sample} (a) shows, MMN adopts the leftmost \emph{pillow} to supervise the first step result (\ie \emph{Select(pillow)}), while our proposed DIS method takes into account all possible correct results. Such comprehensive supervisions alleviate the missing-in-the-middle problem, thus facilitating the model to generate correct answers. Besides, previous IoU-based intermediate supervisions are possible to introduce noise signals due to the bounding-box overlap problem, which induces the model to focus on irrelevant objects. This is exemplified in Figure \ref{fig:sample} (b). In the first reasoning step, the model is required to focus on the \emph{napkin} (\ie probability distribution corresponding to the green lines). However, due to the overlap between \emph{napkin} and other objects (\eg \emph{table}, \emph{laptop}, \emph{cup}, \etc), the IoU-based supervisions prompt the model to focus on irrelevant areas (\ie probability distribution corresponding to the red lines), thus leading to grounding ambiguity and generating wrong answers. 

To this end, we propose a novel Detection-based Intermediate Supervision (DIS) method to resolve aforementioned issues. Specifically, to obtain comprehensive intermediate results, executive programs are parsed from the questions, and then step-by-step inference is performed on the scene graphs, which generates complete intermediate results (illustrated in Figure \ref{fig:sample} (a)). Afterwards, a generative framework is proposed to supervise VQA model using the intermediate results, which transforms the intermediate supervisions into sequences, and constrains model states via sequence generation (illustrated in Figure \ref{fig:sample} (b)). Compared to previous methods, our proposed DIS provides more comprehensive supervision signals for the reasoning process, and exploits a unified generative framework to constrain the intermediate states, thereby improving the answer prediction performance. Moreover, due to the consideration of intermediate results, the answering consistency among compositional questions and their sub-questions is significantly improved.
In summary, the main contributions are the following,
\begin{itemize}
    \item We introduce Detection-based Intermediate Supervision (DIS), a novel method that provides more comprehensive intermediate supervisions via a unified generative framework. To the best of our knowledge, this is the first attempt in the usage of a generative framework for intermediate supervisions in visual question answering.
    \item We propose a scene graph inference framework, which step-by-step executes programs parsed from questions on scene graphs to obtain intermediate results. The supervision signals are further constructed by converting the results into a unified sequential form.
    \item We conduct extensive experiments to evaluate the effectiveness of our proposed DIS algorithm, in which our method achieves competitive answer prediction performance (61.31\% \emph{vs.} 60.83\%), and superior reasoning consistency (73.11\% \emph{vs.} 71.47\%, 64.20\% \emph{vs.} 61.94\%, and 55.28\% \emph{vs.} 52.80\%) compared to previous methods. 
\end{itemize}

% \vspace{-0.3cm}

\section{Related Work}
\subsection{Visual Question Answering}
% \vspace{-0.2cm}
Compositional visual question answering task is defined to generate answers for given compositional questions based on the image content. Generally, compositional questions consist of multiple visual concepts (\ie objects, attributes, and relations), and require VQA models to perform multi-hop reasoning to get the right answers. Recently, several attempts have been made to facilitate visual and logical reasoning, and these methods can be divided into two categories: holistic and modular. Holistic methods \cite{Anderson2017BottomUpAT, Kim2018BilinearAN, Tan2019LXMERTLC,Hu2019LanguageConditionedGN, Yu2019DeepMC, Yang2020TRRNetTR} exploit a unified multimodal fusion model to solve all types of question, and achieve implicit visual/logical reasoning through graph structures \cite{Li2019RelationAwareGA,Hu2019LanguageConditionedGN} and relational attention mechanisms\cite{Li2019RelationAwareGA,Hu2019LanguageConditionedGN,Yang2020TRRNetTR}. With the help of scene graph structure, images are represented as graphical webs, containing information about objects, attributes and relationships among interconnected objects, which can be used for visual reasoning via graph traversal. For example, NSM \cite{Hudson2019LearningBA} performs sequential reasoning over the probabilistic scene graph of the image, and achieves multi-hop inference by shifting probability distributions. RPR \cite{Jing2022LearningTD} casts visual reasoning as a path routing task, and adopts reinforcement learning to explore the inference path. 

\begin{figure*}[!ht]
    \centering
    \includegraphics[width=\textwidth]{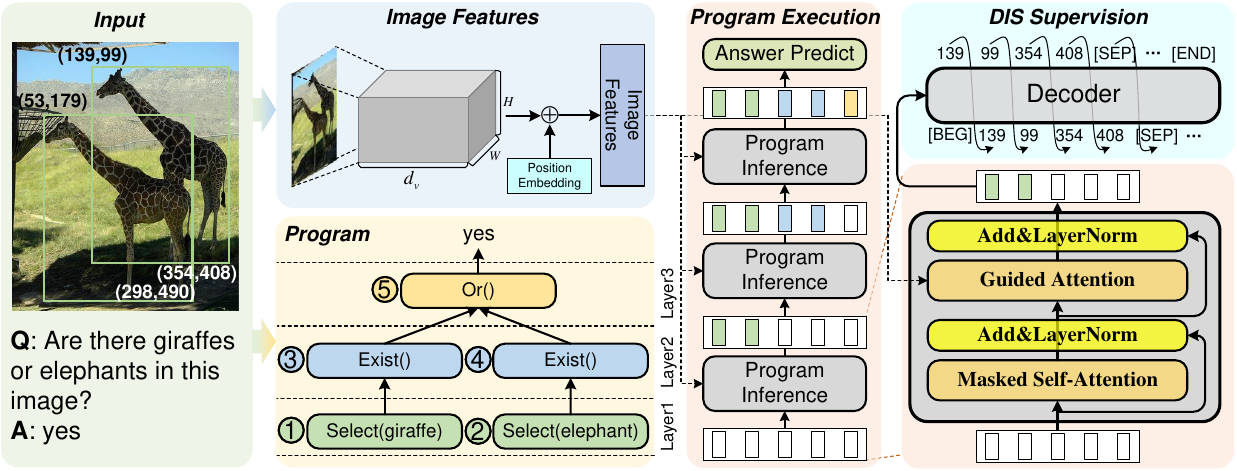}
    \caption{The framework of our proposed Detection-based Intermediate Supervision (DIS) method. Questions are parsed into programs, and step-by-step program execution is performed to generate answers. The intermediate states are supervised by our DIS algorithm.}
    \label{fig:model}
    \vspace{-0.5cm}
\end{figure*}

Despite the overwhelming successes achieved, holistic models process all types of questions via a unified model, which ignores the reasoning structure implicit in the question. Therefore, modular methods \cite{Andreas2015NeuralMN,Hu2018ExplainableNC,Hudson2018CompositionalAN,Chen2019MetaMN,Zhao2021ProToPT} are proposed to make up for the above-mentioned issues. Specifically, modular methods parse the question into a structured tree that reflects the reasoning process, and construct the question-specific model using pre-defined modules. Due to the explicit reasoning structure, such methods have strong interpretability and controllability. In addition, extra intermediate supervisions can be provided to constrain models to reason along prescribed directions, \eg IoU-based Kullback-Leible (KL) divergence \cite{Chen2019MetaMN,Zhao2021ProToPT}, thereby improving answer prediction performance. However, such IoU-based supervisions suffer from two issues, \ie ignorance of multiple grounded objects and grounding ambiguity, underutilizing the intermediate supervisions for model optimization.

\subsection{Object Detection}
There has been a tremendous amount of work in object detection tasks, which require extracting objects from images. Traditional object detection algorithms introduce explicit prior knowledge via producing a set of proposals \cite{Girshick2015FastR,Ren2015FasterRT}, anchors \cite{Redmon2015YouOL}, or window centers \cite{Tian2019FCOSFC}, and then perform non-maximum suppression \cite{Bodla2017SoftNMSI} to remove duplicate predictions. To avoid complex processing procedures, DETR \cite{Carion2020EndtoEndOD} exploits Transformer-based encoder-decoder framework \cite{Vaswani2017AttentionIA} for object detection, which learns a set of ``object queries'' to directly generate bounding boxes and object labels. All of these detectors require extra modules for bounding box regression and label prediction to obtain final predictions. To further avoid such complexities, Pix2Seq \cite{Chen2021Pix2seqAL} directly predicts the raw pixel coordinates through an encoder-decoder network, which achieves competitive performance while simplifying the detection framework. Inspired from Pix2Seq and language modeling \cite{Brown2020LanguageMA}, we propose a detection-based supervision framework, which converts the intermediate results into a unified sequential form that consists of pixels and tokens, thereby providing more comprehensive supervision signals.

\section{Methodology}
We propose Detection-based Intermediate Supervision (DIS) methods to facilitate the constraint of intermediate reasoning state, thereby improving the answer prediction performance. The overall framework is depicted in Figure \ref{fig:model}. Specifically, image features are first extracted via convolutional neural networks (CNN). Then, the question is parsed into program tree, followed by the program execution network to get the final answer. Finally, we introduce detection-based intermediate supervision framework to enhance reasoning ability of the VQA model.

\subsection{Image Features Extraction}
\label{sec:image_feature_extraction}

Image feature extraction is based on the pre-trained Faster R-CNN model \cite{Ren2015FasterRT,Jiang2020InDO}. In contrast to previous methods \cite{Anderson2017BottomUpAT,Hu2019LanguageConditionedGN} that adopt bottom-up features, we adopt the feature maps output from $C5$ layer of Faster R-CNN \cite{Jiang2020InDO} for image representation. Specifically, given an image $I$, the pre-trained CNN backbone of Faster R-CNN is utilized to extract the feature map $\matrix{V}\in\mathbb{R}^{HW\times{d_v}}$, where $H,W$ indicate height and width of the feature map, respectively. $d_v$ denotes the feature dimension. 

To endow the image features with visual contexts and cross-modal textual information, we follow MCAN \cite{Yu2019DeepMC} method, which adopts Transformer block to encode the question and image. Specifically, given a question $Q$ of length $T$, which is embedded into latent space $\matrix{E}\in\mathbb{R}^{T\times{d_h}}$, a two-layer Transformer is adopted to encode the questions as follows:
\begin{equation}
% \begin{split}
\begin{aligned}
    \hat{E}&=\textrm{LN}(\matrix{E}+\textrm{SA}(\matrix{E}))\\
    \tilde{\matrix{E}}&=\textrm{LN}(\hat{\matrix{E}}+\textrm{FFN}(\hat{\matrix{E}})),
\end{aligned}
% \end{split}
\end{equation}
where $\textrm{SA}, \textrm{LN}, \textrm{FFN}$ denote self-attention, layer normalization and feed forward network, respectively. Afterwards, the image feature $\matrix{V}$ is enriched via a two-layer Transformer using visual contexts and question semantics $\matrix{\tilde{E}}$ as follows:
\begin{equation}
\begin{aligned}
    \matrix{V'}&=\textrm{FC}(\matrix{V}+\textrm{PosEmb})\\
    \bar{\matrix{V}}&=\textrm{LN}(\matrix{V'}+\textrm{SA}(\matrix{V'})) \\
    \hat{\matrix{V}}&=\textrm{LN}(\bar{\matrix{V}}+\textrm{GA}(\bar{\matrix{V}},\hat{\matrix{E}})) \\
    \tilde{\matrix{V}}&=\textrm{LN}(\hat{\matrix{V}}+\textrm{FFN}(\hat{\matrix{V}})),
\end{aligned}
\end{equation}
where $\textrm{PosEmb}$ indicates position embedding. $\textrm{FC}$ denotes fully-connected layer, converting feature dimension from $d_v$ to $d_h$. $\textrm{GA}$ denotes guided attention, which exploits question semantics to enhance the relevant visual features. The resulting image representation $\tilde{\matrix{V}}$ can be used for program execution to get the final answer.

\subsection{Program Generation}
\label{sec:program_generation}

Program generation aims to parse questions into program trees that reflect the reasoning procedures inferred by questions, and the program trees can be further used for model construction. We follow MMN \cite{Chen2019MetaMN} to generate programs from questions. Specifically, the nodes of the program tree are formalized as ``Function(Arg1,...ArgN)'', where ``Function'' can be categorized into 10 different abstract types (\eg select, relate, exist, or, \etc), and each abstract type is further subdivided into more subtypes (\eg relate: relate\_attr, relate\_name, relate\_inv\_name, \etc), which take a variable number of arguments as inputs.

\begin{figure}[t]
    \centering
    \includegraphics[width=\linewidth]{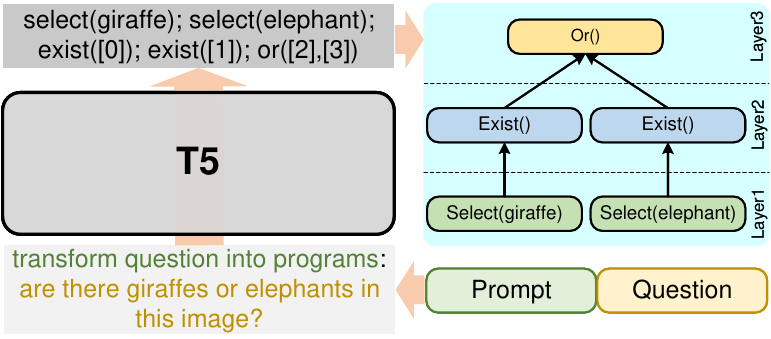}
    \caption{Illustration of the program generation based on encoder-decoder network.}
    \label{fig:program_generator}
    \vspace{-0.4cm}
\end{figure}

Based on the abovementioned program types, the complete program tree can be viewed as the sequence of functional nodes, and generated from an encoder-decoder network, \eg T5 \cite{Raffel2019ExploringTL}. As illustrated in Figure \ref{fig:program_generator}, a prompt (\ie ``transform question into programs:'') is added in front of the question, and fed to T5 model for program generation. The output sequence consists of inter-dependent functions, where ``[N]'' denotes the dependencies. Afterwards, a structured program tree with $L$ layers is constructed from the sequence according to the dependencies, which can be used to guide step-by-step program execution.

\subsection{Program Execution}
\label{sec:program_execution}
Given image features $\tilde{\matrix{V}}$ and the $L$-layer program tree, program execution performs step-by-step inference from Layer-$1$ to Layer-$L$ based on image features to get the answer, and all layers share model parameters, making it parameter-efficient and scalable to any number of layers. Specifically, suppose the program tree contains $N$ nodes, and each node corresponds to a program text (\eg ``Select(giraffe)'', ``Select(elephant)'', \etc), a state matrix $\matrix{S}\in\mathbb{R}^{N\times{d_h}}$ is initialized with the program semantics as follows:
\begin{gather}
\begin{aligned}
    \{\vec{e}_0^i,\vec{e}_1^i,...,\vec{e}_k^i\}&=\textrm{GloVe}(prog_i) \\
    \matrix{S}_i&=\textrm{FC}(\textrm{Concat}(\{\vec{e}_0^i,\vec{e}_1^i,...,\vec{e}_l^i\})),
\end{aligned}
\end{gather}
where $prog_i$ denotes the program text of $i$-th node, and is truncated or padded to fixed length $k$. $\textrm{GloVe}$ denotes the GloVe embedding layer \cite{Pennington2014GloVeGV}. In the process of step-by-step inference, the matrix $\matrix{S}$ implicitly contains the intermediate reasoning states and results, and can be used to decode outputs for our proposed DIS algorithm.

Denote the input state of the $l$-th layer as $\matrix{S}^{l-1}$, we exploit Transformer framework to obtain the output state $\matrix{S}^{l}$. Specifically, a masked self-attention layer is firstly exploited to gather dependencies from the state of last layer, and then a guided attention layer is utilized to find visual clues from image features $\tilde{\matrix{V}}$, formulated as follows:
\begin{gather}
\begin{aligned}
    \hat{\matrix{S}}^{l-1}&=\textrm{LN}(\matrix{S}^{l-1}+\textrm{MaskSA}(\matrix{S}^{l-1},\matrix{M}^{l})) \\
    \matrix{S}^{l}&=\textrm{LN}(\hat{\matrix{S}}^{l-1}+\textrm{GA}(\hat{\matrix{S}}^{l-1},\tilde{\matrix{V}})),
\end{aligned}
\end{gather}
where $\textrm{MaskSA},\textrm{GA}$ denote mask self-attention and guided attention layer, respectively. $\textrm{MaskSA}$ uses weight matrix $\matrix{M}^l$ to mask non-dependent nodes, formulated as follows:
\begin{gather}
\begin{aligned}
    \textrm{MaskSA}(\matrix{S},\matrix{M})&=\textrm{Softmax}(\frac{(\matrix{Q}^S)(\matrix{K}^S)^T}{\sqrt{d_h}}+\matrix{M})\matrix{V}^S,
\end{aligned}
\end{gather}
where $\matrix{M}\in\mathbb{R}^{N\times{N}}$ denotes the mask matrix. $\matrix{M}_{ij}=0$ if and only if $i$-th node is the parent node of $j$-th node, and otherwise $\matrix{M}_{ij}=-\infty$. $\matrix{Q}^S,\matrix{K}^S,\matrix{V}^S$ are derived from $\matrix{S}$ via three fully-connected layers, formulated as follows:
\begin{gather}
\label{eq:sa_qkv}
\begin{aligned}
    \matrix{Q}^S&=\textrm{FC}^Q(\matrix{S}) \\
    \matrix{K}^S&=\textrm{FC}^K(\matrix{S}) \\
    \matrix{V}^S&=\textrm{FC}^V(\matrix{S}). \\
\end{aligned}
\end{gather}

Different from $\textrm{SA}$ that gathers information from itself, guided attention gathers features from other source (\eg visual clues), formulated as follows:
\begin{gather}
    \textrm{GA}(\matrix{S},\tilde{\matrix{V}})=(\frac{(\matrix{Q}^S)(\matrix{K}^{\tilde{V}})^T}{\sqrt{d_h}})\matrix{V}^{\tilde{V}},
\end{gather}
where $\matrix{K}^{\tilde{V}},\matrix{V}^{\tilde{V}}$ are derived from $\tilde{\matrix{V}}$ similar to Equation \ref{eq:sa_qkv}.

After $L$ iterations of program inference, the final state $\matrix{S}^L_{N-1}$ is used to predict the answer via a 
multi-layer perception (MLP) layer:
\begin{gather}
\begin{aligned}
    \vec{s}&=\textrm{MLP}(\matrix{S}^L_{N-1}) \\
    p(a|I,Q;\Theta)&=\textrm{Softmax}(\vec{s}),
\end{aligned}
\end{gather}
where $\vec{s}\in\mathbb{R}^{|\mathcal{A}|}$ denotes the predicted scores of the answers in $\mathcal{A}$, and the answer with the highest score is chosen as the final answer. $\Theta$ denotes the model parameters. Finally, cross-entropy loss is used to optimize the model, formulated as follows:
\begin{gather}
    \label{vqa_loss}
    \mathcal{L}^{VQA}=-\mathbb{E}_{\mathcal{D}}[\log(p(a=a^{gt}|I,Q;\Theta))],
\end{gather}
where $\mathcal{D},a^{gt}$ denotes the VQA dataset and ground truth answer, respectively.

% \subsection{Detection-based Intermediate Supervision}
\subsection{Intermediate Supervision}
\label{sec:dis}

\begin{figure}[t]
    \centering
    \includegraphics[width=\linewidth]{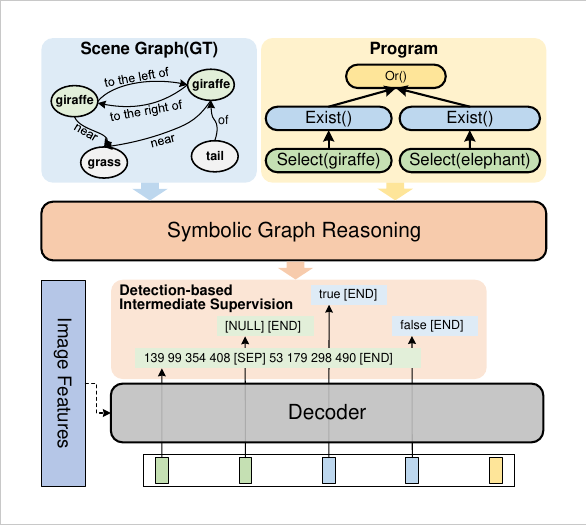}
    \caption{Illustrations of Detection-based Intermediate Supervision (DIS) algorithm. The intermediate results are generated via symbolic graph reasoning based on the ground-truth scene graph and program tree, and transformed into sequences to supervise the intermediate states.}
    \label{fig:dis}
    \vspace{-0.4cm}
\end{figure}
On top of the above-mentioned program execution, intermediate supervisions are proposed to constrain the reasoning process, and improve the answer prediction performance \cite{Chen2019MetaMN,Zhao2021ProToPT}. However, previous methods calculate probability distributions based on IoU between object bounding boxes and ground-truth ones, which easily induces the model to focus on irrelevant objects (ref to Figure \ref{fig:sample} (b)). To this end, we propose detection-based intermediate supervision (DIS) algorithm, which formulates the intermediate supervisions into a unified sequence form, thereby endowing the model with the abilities of exploiting diverse supervision types (\eg bounding boxes, logical words (true/false), text). 
As shown in Figure \ref{fig:dis}, DIS algorithm consists of two steps: (1) \emph{symbolic graph reasoning} designs manual rules to execute the program tree on the ground-truth scene graph\footnote{GQA \cite{Hudson2019GQAAN} dataset provides ground-truth scene graph only for \emph{train} and \emph{val} splits. Therefore, DIS is used to optimize the model only in the training phase, and removed during testing phase.}, resulting in intermediate results (\eg \emph{Objects}, \emph{True/False}, \emph{Answer}), and (2) \emph{intermediate result decoder} decodes the intermediate outputs, which is optimized using auto-regression loss.

\subsubsection{Symbolic Graph Reasoning}
\label{sec:symbolic_graph_reasoning}
To obtain the intermediate supervisions, we follow MMN \cite{Chen2019MetaMN}, which executes the \emph{Program} on the ground truth \emph{Scene Graph} (as illustrated at the top of Figure \ref{fig:dis}). Specifically, we manually design the symbolic reasoning rules for each function type (\eg Select, Relate, Verify, \etc). For example, \emph{Select(x)} is defined to \emph{select the nodes corresponding to x}, and \emph{Relate(x, to the left of)} is defined to \emph{find the nodes to the left of x}. With the \emph{Program} and \emph{Scene Graph}, the intermediate supervisions are inferred step by step, as illustrated at the bottom of Figure \ref{fig:dis}. 
Generally, the supervisions are categorized into three types: \emph{Objects} with bounding boxes, \emph{True/False}, and \emph{Answers}. Regarding \emph{Objects}, we follow Pix2Seq \cite{Chen2021Pix2seqAL} to quantize coordinates into bins, which can be regarded as discrete labels. If multiple objects exist, their bounding boxes are randomly shuffled and concatenated to form the result sequence. For \emph{True/False} and \emph{Answer}, we directly use the textual tokens to form the sequence. Additionally, several special tokens are added to sequence for further generation, \eg [BEG], [SEP], [END], \etc

\subsubsection{Intermediate Result Decoder} With the help of intermediate results, VQA model can be optimized using these supervisions. Specifically, a two-layer Transformer is proposed to decode the intermediate outputs from the states $\tilde{\matrix{S}}^L$. Firstly, [BEG] token is initialized with the state $\tilde{\matrix{S}}^L_i$, and prompts the decoder to generate output sequence. In the training phase, we use teacher-forcing and auto-regression loss to optimize the model, formulated as follows:
\begin{gather}
\label{dis_loss}
\begin{aligned}
    \{y_0, y_1, ..., y_{o-1}\}&=\textrm{Decoder}(\tilde{\matrix{S}}^L_i,\tilde{\matrix{V}}) \\
    \mathcal{L}^{DIS}&=-\frac{1}{o}\sum_{i=0}^{o-1}\log(p(y_i|y_0,...,y_{i-1},\tilde{\matrix{S}}^L_i,\tilde{\matrix{V}})),
\end{aligned}
\end{gather}
where $o$ denotes the length of the result sequence. $\tilde{\matrix{S}}^L_i\in\mathbb{R}^{d_h}$
denotes the intermediate state from the $i$-th program node.

\subsubsection{Model Optimization}
In the training phase, the answer prediction loss (Equation \ref{vqa_loss}) and detection-based intermediate supervision loss (Equation \ref{dis_loss}) are combined to optimize the model:
\begin{gather}
\mathcal{L}=\mathcal{L}^{VQA}+\alpha\mathcal{L}^{DIS},
\end{gather}
where $\alpha$ denotes the loss weight of DIS. In the testing phase, DIS module can be removed because only the final answer needs to be predicted.

\vspace{-0.05cm}
\section{Experiments}
% Datasets
% Implementation Details
% Experimental Results
% Ablation Studies
% Visualization

\subsection{Datasets}

To evaluate the answer prediction performance and answering consistency, the reported results in the following sections are evaluated on the widely used GQA \cite{Hudson2019GQAAN} dataset, and its variant GQA-Sub \cite{Jing2022MaintainingRC}.
\textbf{GQA} \cite{Hudson2019GQAAN} is a compositional visual question-answer dataset, which features compositional questions over the real-world images. It is designed to provide accurate indications of visual understanding capacities and mitigate the language priors that exist widely in previous VQA datasets \cite{Agrawal2015VQAVQ}. 
\textbf{GQA-Sub} \cite{Jing2022MaintainingRC} is derived from the well-organized GQA dataset, and creates sub-questions for \emph{train} and \emph{val} splits, thereby enabling quantitative evaluation of reasoning consistency. More information about GQA and GQA-Sub, as well as their respective evaluation metrics, can be found in Appendix-A.

\subsection{Implementation Details}
\vspace{-0.1cm}
\paratitle{Program Generation.} T5-base\footnote{\url{https://huggingface.co/docs/transformers/model_doc/t5}} is utilized for text transformation, where source and target texts are limited to a maximum length of 40 and 100, respectively. We exploit AdamW optimizer with learning of $1e^{-4}$ and batch size of 32 to finetune T5 for 400k steps. 

\paratitle{Visual Question Answering.} Following the settings from MMN \cite{Chen2019MetaMN}, the questions are truncated or padding to the fixed length of 32. The number of nodes in each program tree is limited to 9, and the maximum length $k$ of each program is set to 8. See the Appendix-B for more implementation details.

% Table generated by Excel2LaTeX from sheet 'GQAResults'
\begin{table}[t]
\setlength{\tabcolsep}{1mm}
    \centering
    \begin{tabular}{llccc}
    \toprule
    \textbf{Model} & \textbf{Required Inputs} & \textbf{Binary} & \textbf{Open} & \textbf{Accuracy} \\
    \midrule
    BUTD  & V+L   & 66.64  & 34.83  & 49.74  \\
    MAC   & V+L   & 71.23  & 38.91  & 54.06  \\
    GRN   & V+L   & 74.93  & 41.24  & 57.04  \\
    LCGN  & V+L   & 73.77  & 42.33  & 57.07  \\
    BAN   & V+L   & 76.00  & 40.41  & 57.10  \\
    PVR   & V+L+Program & 74.58  & 42.10  & 57.33  \\
    LXMERT & V+L+Pre-training & 77.16  & \textbf{45.47}  & 60.33  \\
    % NSM   & V+L+SceneModel & \textbf{78.94 } & \textbf{49.25 } & \textbf{63.17 } \\
    \midrule
    MCAN  & V+L   & 75.87  & 42.15  & 57.96  \\
    RPR   & V+L   & -     & -     & 59.43  \\
    RCVQA & V+L+DataAug & -     & -     & 59.58  \\
    MMN   & V+L+Program & 78.90  & 44.89  & 60.83  \\
    DIS   & V+L+Program & \textbf{79.36 } & 45.36  & \textbf{61.31 } \\
    \bottomrule
    \end{tabular}%
    \vspace{-0.2cm}
    \caption{Comparison of DIS single model with the state-of-the-art methods on the blind test2019 leaderboard of GQA.}
\label{tab:gqa_result}%
\vspace{-0.45cm}
\end{table}% 

\paratitle{Baselines.} Our model is compared with various state-of-the-art approaches excerpted from MMN, including BUTD \cite{Anderson2017BottomUpAT}, MAC \cite{Hudson2018CompositionalAN}, GRN \cite{Guo2019GraphRN}, LCGN \cite{Hu2019LanguageConditionedGN}, BAN\cite{Kim2018BilinearAN}, PVR \cite{Li2019PerceptualVR}, LXMERT \cite{Tan2019LXMERTLC}, MCAN \cite{Yu2019DeepMC}, MMN \cite{Chen2019MetaMN}, RPR \cite{Jing2022LearningTD}, and RCVQA \cite{Jing2022MaintainingRC}. We did not compare with the NSM \cite{Hudson2019LearningBA} because it utilizes a well-tuned external scene graph generation model. 
More information about baselines are provided in Appendix-C.

\vspace{-0.2cm}
\subsection{Experimental Results}
In the experiments, we primarily assessed the performance of our model on answering compositional questions as well as its performance in reasoning consistency. The corresponding experimental results are presented in Table \ref{tab:gqa_result} and Table \ref{tab:gqa_rc}, respectively.
The online test results of the state-of-the-art models and our proposed DIS method on the GQA dataset are shown in Table \ref{tab:gqa_result},
and these results also reflect the performance of all models on compositional questions.
The required inputs represent the information necessary for the model to predict the answers, where V and L indicate vision and language, respectively, while DataAug represents data augmentation.
% As Table \ref{tab:gqa_result} shows, 
% our proposed DIS approach achieves competitive results compared to previous methods. 
% Specifically, NSM exploits a well-designed scene graph generation model, which provides structured prior information, greatly facilitating the visual reasoning and achieves relatively higher accuracies (63.17\% overall accuracy), especially for \emph{Open} questions (49.25\% \emph{Open} accuracy). However, it can be observed that our proposed DIS method performs better on \emph{Binary} questions compared to other scene graph-based models (79.36\% \emph{Binary} accuracy), \eg NSM, LCGN. It might be the reason that the program tree structure is more suitable for logical reasoning, while scene graph does well on questions involving relational reasoning among objects. Excluding the scene graph information, 
As shown, our proposed DIS achieves the best on \emph{Binary}, \emph{Open}, and \emph{Overall} accuracies among the methods listed in Table \ref{tab:gqa_result}. Specifically, with basically the same inputs and settings as MMN, DIS method outperforms MMN by a margin of +0.46\% and +0.47\% for \emph{Binary} and \emph{Open} questions, respectively.

% Table generated by Excel2LaTeX from sheet 'GQAResults'
\begin{table}[t]
    \setlength{\tabcolsep}{1mm}
    \centering
    \begin{tabular}{lccccc}
    \toprule
    \textbf{Methods} & \textbf{Acc} & \textbf{Acc(Sub)} & \textbf{RC(1)} & \textbf{RC(2)} & \textbf{RC(3)} \\
    \midrule
    Language-only & 43.86  & 41.16  & 31.61  & 18.81  & 7.31  \\
    Visual-only & 56.63  & 53.97  & 46.63  & 28.16  & 16.26  \\
    MAC   & 62.08  & 62.63  & 56.10  & 41.67  & 33.96  \\
    MAC + DA & 61.04  & 71.42  & 65.78  & 54.09  & 43.14  \\
    LCGN  & 64.16  & 63.74  & 57.37  & 44.32  & 35.09  \\
    LCGN + DA & 64.14  & 73.46  & 68.93  & 58.94  & 50.05  \\
    MMN   & 65.05  & 64.46  & 58.79  & 43.98  & 33.96  \\
    MMN + DA & 65.65  & 74.60  & 69.59  & 57.98  & 48.17  \\
    RCVQA + DA & 66.26  & 76.02  & 71.47  & 61.94  & 52.80  \\
    \midrule
    DIS   & 69.71  & 70.31  & 64.98  & 53.55  & 48.86  \\
    DIS + DA & \textbf{69.81 } & \textbf{77.91 } & \textbf{73.11 } & \textbf{64.20 } & \textbf{55.28 } \\
    \bottomrule
    \end{tabular}%
    \vspace{-0.1cm}
    \caption{Results of our proposed DIS method and the state-of-the-arts on GQA-Sub. The ``DA'' denotes the data augmentation.}
\label{tab:gqa_rc}%
\vspace{-0.5cm}
\end{table}%  
Also, we evaluate the performance of the proposed DIS in terms of reasoning consistency. The results regarding accuracy (\ie Acc and Acc(Sub)) and reasoning consistency (\ie $\textrm{RC}(k)$, refer to Appendix-A for details) of our proposed DIS and state-of-the-art methods are presented in Table \ref{tab:gqa_rc}. Acc and Acc(Sub) denote the accuracies on \emph{val} and \emph{val-sub} splits, respectively. $\textrm{DA}$ indicates the usage of augmented sub-questions for model training. As Table \ref{tab:gqa_rc} shows, our proposed DIS surpasses the other state-of-the-art methods on both accuracy and reasoning consistency metrics. 
Specifically, without data augmentation of sub-questions, DIS outperforms MMN by a large margin of 4.66\% and 5.85\% on \emph{val} and \emph{val-sub} splits, respectively. The reasoning consistency is also significantly improved by using our DIS algorithm, \ie a margin of \textbf{6.19}\%, \textbf{9.57}\%, and \textbf{14.9}\% on $\textrm{RC}(1)$, $\textrm{RC}(2)$, and $\textrm{RC}(3)$, respectively.
Such superiority stems from the comprehensive yet noise-free supervisions of intermediate results provided by DIS, significantly enhancing the ability of the model to answer compositional questions and corresponding sub-questions. 
When we trained the DIS with sub-questions in \emph{train-sub} as a form of data augmentation, our proposed DIS even outperforms the best-performing RCVQA model that is tailored for enhancing reasoning consistency through the incorporation of a consistency constraint loss. Specifically, DIS surpasses RCVQA by a significant margin of 3.55\% on Acc and over 1.64\%, 2.26\% and 2.48\% on the three reasoning consistency metrics, respectively, which indicates the effectiveness of our proposed method.

\vspace{-0.25cm}
\subsection{Ablation Studies}
\vspace{-0.1cm}
\begin{figure*}[!ht]
    \centering
    \includegraphics[width=\textwidth]{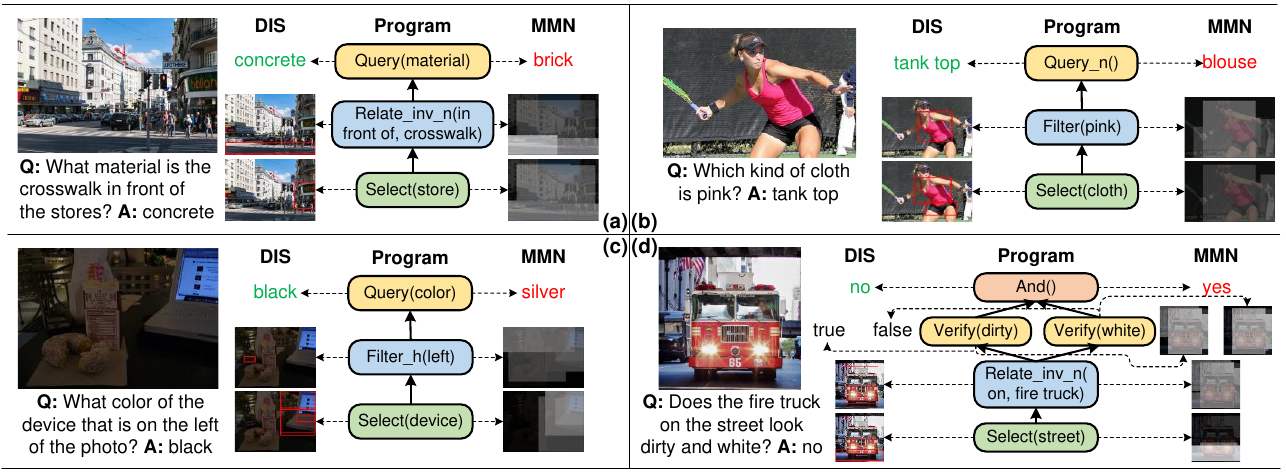}
    \vspace{-0.5cm}
    \caption{Visualization of intermediate results and predicted answers from MMN and our proposed DIS model. }
    \label{fig:cases}
\vspace{-0.4cm}
\end{figure*}
In this section, a series of ablations are conducted on GQA dataset to investigate the effectiveness of our proposed method. All the models are trained on \emph{train+val} split, and evaluated on \emph{testdev} split. The experimental settings are kept consistent throughout ablation studies. 

% Table generated by Excel2LaTeX from sheet '正交实验表格'
\begin{table}[t]
    \centering

    \begin{tabular}{lc}
    \toprule
    \textbf{Format} & \textbf{Accuracy} \\
    \midrule
    name, x1, y1, x2, y2 & 59.40  \\
    x1, y1, x2, y2, name & 59.85  \\
    x1, y1, x2, y2 & \textbf{59.97 } \\
    \bottomrule
    \end{tabular}%
    \caption{Results of different object supervision formats.}
\label{tab:abl_format}%
\vspace{-0.5cm}
\end{table}%

\paratitle{Different \emph{Object} supervision format: }Table \ref{tab:abl_format} shows the results with different object supervision formats. $(x1, y1)$, $(x2, y2)$ and $name$ denote the top-left, bottom-right coordinates of bounding box, and object label, respectively. As Table \ref{tab:abl_format} shows, the bounding box information is enough for intermediate supervision, which achieves the highest score (\ie 59.97\%), and the extra object names decrease the performance. It is conjectured that extra supervisions (\ie object names) enforce the VQA model focus more on intermediate results than final answer prediction, thus decreasing the answer prediction performance.

\paratitle{Different loss weights of DIS and bootstrapping epochs: }Table \ref{tab:abl_loss} shows the results of different DIS loss weights, and different epochs for bootstrapping. As shown, the best accuracy (59.97\%) is achieved when DIS loss weight $\alpha=0.5$, which surpasses the baseline $\alpha=0$ by a significant margin of 1.22\%, demonstrating the effectiveness of our proposed DIS method. In addition, it can be observed that the bootstrapping of 1 epoch has achieved the best accuracy (61.31\%), while decreasing the performance with more bootstrapping epochs. It is conjectured that more training epochs on \emph{all-split} bring biases for the VQA model, which is harmful for further fine-tuning.
\textbf{Additionally}, the results regarding how different quantized bins and maximum number of generated objects and the shuffle mode of object order affect the performance are shown in Appendix-D.

% % Table generated by Excel2LaTeX from sheet '正交实验表格'
% \begin{table}[t]
%     \centering
%     \begin{tabular}{cc}
%     \toprule
%     \textbf{Use Shuffle} & \textbf{Accuracy} \\
%     \midrule
%     \XSolidBrush     & 59.57 \\
%     \Checkmark     & \textbf{59.97} \\
%     \bottomrule
%     \end{tabular}%
%     \caption{Results of different shuffle modes for the intermediate result order}
% \label{tab:abl_shuffle}%
% \end{table}%

% Table generated by Excel2LaTeX from sheet '正交实验表格'
\begin{table}[t]
    \centering
    \begin{tabular}{cc|cc}
    \toprule
    \textbf{$\mathbf{\alpha}$} & \textbf{Accuracy} & \textbf{BS Epoch} & \textbf{Accuracy} \\
    \midrule
    0.00     & 58.75 & 1     & \textbf{61.31} \\
    0.25  & 58.91 & 2     & 61.04 \\
    0.50   & \textbf{59.97} & 3     & 60.95 \\
    0.75  & 59.47 & 4     & 60.61 \\
    1.00     & 59.74 & 5     & 60.59 \\
    \bottomrule
    \end{tabular}%
    \caption{Results of different DIS loss weights and bootstrapping epochs.}
\label{tab:abl_loss}%
\vspace{-0.6cm}
\end{table}%

\vspace{-0.2cm}
\subsection{Visualization}
\vspace{-0.1cm}
We visualize several cases from the GQA \emph{testdev} split, showcasing the intermediate results and predicted answers from MMN and DIS. As depicted in Figure \ref{fig:cases}, it is evident that MMN tends to focus on a large area of the image, \eg the attention maps of the \emph{store}, \emph{cloth}, and \emph{fire truck} in Figure \ref{fig:cases} (a), (b), and (d) are not tightly focused on. It might be the reason that the large bounding boxes are more likely to overlap with others. Consequently, these bounding boxes are frequently used for training, leading the model to prioritize larger areas for IoU-based intermediate supervision methods. On the contrary, our proposed DIS method is able to predict the tight bounding boxes, \eg the \emph{crosswalk}, \emph{store} in Figure \ref{fig:cases} (a), \emph{cloth} in Figure \ref{fig:cases} (b), and \emph{street}, \emph{fire truck} in Figure \ref{fig:cases} (d), which facilitates the model to learn more fine-grained cross-modal alignments and accurate reasoning procedures.
In addition, as depicted in Figure \ref{fig:cases} (b) and (c), there may exist multiple intermediate supervisions. However, MMN does not precisely focus on these areas, \eg the \emph{cloth} in Figure \ref{fig:cases} (b). The incomplete prediction of intermediate results makes it easy for the models to infer incorrect answers. In contrast, our proposed DIS is able to predict multiple object results in one sequence and unify the different result forms (\ie logical true/false and objects) using one single framework.

\section{Conclusion}

We propose the DIS algorithm for compositional visual question answering. Specifically, DIS exploits a unified generative framework to provide intermediate supervisions in a sequential form that provides more fine-grained and accurate supervisions, addressing the issue of supervision ambiguity and promoting cross-modal knowledge alignment. 
We conducted experiments on the GQA and GQA-Sub datasets and the experimental results demonstrate that DIS achieves competitive answer prediction performance and superior reasoning consistency compared to previous state-of-the-arts.

% We propose a detection-based intermediate supervision (DIS) algorithm for compositional visual question answering. Specifically, DIS exploits a unified generative framework to provide intermediate supervisions, in a sequential form, for neural module networks. Thus, DIS adopts finer-grained supervision signals to address the supervision ambiguity caused by the overlapping problem of bounding boxes. Moreover, such sequential form offers a more flexible and easy-to-use strategy for model optimization, enabling the handling of multiple objects and logical true/false supervisions. We conducted experiments on GQA datasets to evaluate the performance of our proposed DIS algorithm. The experimental results demonstrate that DIS achieves competitive answer prediction performance and superior reasoning consistency compared to previous state-of-the-arts.

\section{Acknowledgments}
This work was supported in part by the National Natural Science Foundation of China under Grant No. 62276110, No. 62172039 and in part by the fund of Joint Laboratory of HUST and Pingan Property \& Casualty Research (HPL). The authors would also like to thank the anonymous reviewers for their comments on improving the quality of this paper.

% \bibliography{plain}
\bibliography{aaai24}

\clearpage
\appendix
% \noindent This part is the Appendix of the main paper. The detail information about the GQA and GQA-Sub datasets, as well as their respective evaluation metrics, is introduced in Section A, and the implementation details regarding the visual question answering model are shown in Section B. Moreover, the baselines and ablation studies are shown in Section C and Section D, respectively.

\noindent This part serves as the appendix of the main paper. Section A provides detailed information about the GQA and GQA-Sub datasets, along with their respective evaluation metrics. Implementation details of the visual question answering model are presented in Section B. Additionally, Section C covers baselines, while Section D introduces more ablation studies.

\section{A. Datasets}
\label{datasets}
\paratitle{GQA}: Specifically, GQA \cite{Hudson2019GQAAN} is generated by leveraging Visual Genome \cite{Krishna2016VisualGC} scene graphs to create diverse reasoning questions with less language bias. Therefore, it requires more complicated reasoning capacities to answer the questions. GQA consists of two splits (\ie \textit{balance-split} and \textit{all-split}). The \emph{balance-split} consists of QA pairs with re-sampled question-answer distribution. Following the common practice \cite{Chen2019MetaMN}, we use \emph{all-split} for bootstrapping, and \emph{balance-split} for finetuning and online evaluation. The dataset is split into 70\% train, 10\% validation, 10\% test and 10\% challenge. The metrics utilized in this paper include accuracy regarding the \textit{Open} questions, \textit{Binary} questions and the overall accuracy. 

\paratitle{GQA-Sub}:
Specifically, GQA-Sub \cite{Jing2022MaintainingRC} transforms compositional questions into language graphs and extracts sub-graphs to construct sub-questions. To obtain the answer labels for these questions, GQA-Sub exploits ground-truth scene graphs and language graph traversing to get the answers. Additionally, to avoid the language biases, GQA-Sub performs three times of sampling for these generated samples, resulting in 351,271 and 45,043 sub-questions for \emph{train} and \emph{val} splits, respectively. Finally, a total of 4 splits are generated, \ie \emph{train}, \emph{train-sub}, \emph{val}, and \emph{val-sub}, where the \emph{train/val-sub} split contains sub-questions corresponding to \emph{train/val} split. Following the common settings \cite{Jing2022MaintainingRC}, the VQA model is trained on \emph{train(-sub)}, and evaluated on \emph{val(-sub)}. 

To measure the reasoning consistency of VQA models, GQA-Sub designes a metric, \ie reasoning consistency score $\textrm{RC}(k)$, computed by:
\begin{gather}
    \textrm{RC}(k)=\frac{\sum_{Q\subset\mathbb{Q}, N\ge{k}}\textrm{Correct}^f(Q,\{Q_i\}_{i=1}^{N})}{\sum_{Q\subset\mathbb{Q}, N\ge{k}}\textrm{Correct}^f(Q)}
\end{gather}
where $\mathbb{Q}$ denotes the set of compositional questions. $\textrm{Correct}(q)=1$ only when $q$ is correctly answered by the VQA model, or $\textrm{Correct}(q)=0$ otherwise. $\textrm{Correct}^f(Q,\{Q_i\}_{i=1}^{N})=1$ only when the compositional question $Q$ and all of its sub-questions $\{Q_i\}_{i=1}^{N}$ are correctly answered, or $\textrm{Correct}^f(Q,\{Q_i\}_{i=1}^{N})=0$ otherwise. The value of $\textrm{RC}(k)$ ranges from 0 to 1, and indicates better consistency when $\textrm{RC}(k)$ is higher.

\section{B. Implementation Details}

\paratitle{Visual Question Answering.} As for program execution, the number of reasoning layers $L$ is set to 5. In the process of intermediate output decoding, the maximum number of generated objects is set to 4. The bins for quantizing coordinates are set to 256. Regarding model training, the loss weight of DIS $\alpha$ is 0.5. Following the settings from MMN \cite{Chen2019MetaMN}, the VQA model is firstly bootstrapped for 5 epochs on \emph{all-split}, and then finetuned on \emph{balance-split}. In the bootstrapping stage, the model is trained using Adam optimizer with a fixed learning rate of $1e^{-4}$ and batch size of 128. In the fine-tuning stage, the model is optimized using Adam optimizer with batch size of 128 and base learning rate of $2e^{-4}$. In addition, warmup \cite{Goyal2017AccurateLM} strategy is exploited to facilitate model fine-tuning. Specifically, the learning rate increases linearly from $1e^{-4}$ to $2e^{-4}$ for the first 4 epochs, and decays by 0.5 for every 2 epochs at epoch 10. The model is trained for 18 epochs in total. 

\section{C. Baselines}
The baselines include BUTD \cite{Anderson2017BottomUpAT}, MAC \cite{Hudson2018CompositionalAN}, GRN \cite{Guo2019GraphRN}, LCGN \cite{Hu2019LanguageConditionedGN}, BAN\cite{Kim2018BilinearAN}, PVR \cite{Li2019PerceptualVR}, LXMERT \cite{Tan2019LXMERTLC}, 
% NSM \cite{Hudson2019LearningBA}, 
MCAN \cite{Yu2019DeepMC}, MMN \cite{Chen2019MetaMN}, RPR \cite{Jing2022LearningTD}, and RCVQA \cite{Jing2022MaintainingRC}. A brief introduction of these baseline methods is described as follows.
\begin{itemize}
    \item \textbf{BUTD.} The classic attention-based model for visual question answering and image caption, which exploits top-down attention to capture the question-relevant visual features for answer prediction.
    \item \textbf{MAC.} The memory-based reasoning model, which decomposes the question into a series of attention-based reasoning steps, and performs control and memory iteratively to get the answer.
    \item \textbf{GRN.} The graph-based reasoning model, which considers the relationships between words in the question, and designs intra- and inter-modal graph to exchange information from multi-modal inputs, thereby realizing implicit multi-step reasoning.
    \item \textbf{LCGN.} The graph-based reasoning model, which constructs fully-connected object graph, and exploits graph attention network (GAT) to perform implicit visual reasoning based on the graph structure.
    \item \textbf{BAN.} The bilinear attention network, which calculates bilinear attention distributions via low-rank bilinear pooling to facilitate interactions between multimodal inputs. 
    \item \textbf{PVR.} The module-based approach that incorporates the concepts of logical and/or for logical inference, and introduces more perceptual modules for better logical generalization.
    \item \textbf{LXMERT.} The multi-modal pretrained model, which first pretrains transform-based models on large-scale image-text corpus to learn task-agnostic representations, and then finetunes the models on downstream VQA task.
    % \item \textbf{NSM.} The graph-based symbolic reasoning model, which constructs a probabilistic graph for the representation of the representation and performs iterative traversing the graph to answer the question. 
    \item \textbf{MCAN.} The transformer-based multi-modal fusion model, which interacts intra- and inter-modal features using multi-head self-attention and guided attention, resulting in multi-modal joint representations for visual question answering.
    \item \textbf{MMN.} The module-based network, which decomposes questions into an interdependent program sequence, and assembles parametric modules to construct reasoning model for answer generation. 
    \item \textbf{RPR.} The graph-based reasoning model, which formulates the reasoning process as a reinforced path routing problem, and exploits reinforcement learning to optimize the reasoning process. 
    \item \textbf{RCVQA.} The VQA model which focuses on the answering consistency problem and designs a consistency constraint loss to improve the answering consistency between a compositional question and its sub-questions.
\end{itemize}

\section{D. Ablation Studies}

\paratitle{Different quantized bins and maximum number of generated objects: } Table \ref{tab:abl_bins} shows the results when quantizing coordinates into different bins and generating different number of intermediate results. As Table \ref{tab:abl_bins} shows, the highest accuracy (\ie 59.97\%) is achieved when quantized bins are set to $256$ and the maximum number of generated objects is set to $4$. Intuitively, the quantized bins are limited by the size of feature maps. The reason why the accuracy does not increase when \# bins $>{256}$ might be that the feature map with size of $10\times{10}$ is unable to predict such fine-grained coordinates. From Table \ref{tab:abl_bins}, it can also be observed that generating $1\sim{3}$ intermediate results decreases the VQA performance, indicating that more intermediate supervisions are helpful for the model to learn visual reasoning and achieve higher performance.

\begin{table}[!h]
    \centering
    \begin{tabular}{cc|cc}
    \toprule
    \textbf{Bins} & \textbf{Accuracy} & \textbf{Number} & \textbf{Accuracy} \\
    \midrule
    64    & 59.35 & 1     & 59.16 \\
    128   & 59.63 & 2     & 59.56 \\
    256   & \textbf{59.97} & 3     & 59.78 \\
    384   & 59.44 & 4     & \textbf{59.97} \\
    512   & 59.66 & 5     & 59.25 \\
    \bottomrule
    \end{tabular}%
    \caption{Results of different quantized bins and maximum number of generated objects for intermediate supervisions.}
\label{tab:abl_bins}%
\end{table}%

% Table generated by Excel2LaTeX from sheet '正交实验表格'
\begin{table}[!h]
    \centering
    \begin{tabular}{cc}
    \toprule
    \textbf{Use Shuffle} & \textbf{Accuracy} \\
    \midrule
    \XSolidBrush     & 59.57 \\
    \Checkmark     & \textbf{59.97} \\
    \bottomrule
    \end{tabular}%
    \caption{Results of different shuffle modes for the intermediate result order.}
\label{tab:abl_shuffle}%
\end{table}%

\paratitle{Shuffle mode of object order: } Table \ref{tab:abl_shuffle} shows the results of different shuffle modes when providing intermediate results. As illustrated in Figure 1(a), there might exist multiple results during program inference. Therefore, if we do not shuffle the order of the intermediate results, only part of the objects are used for supervision, hindering the model to learn comprehensive reasoning process. As expected, using the shuffle strategy improves the VQA performance (\ie 59.97\% \emph{vs} 59.57\%).

\end{document}